\documentclass[letterpaper]{article} 
\usepackage{aaai2026}
\usepackage{times}  
\usepackage{helvet}  
\usepackage{courier}  
\usepackage[hyphens]{url}  
\usepackage{graphicx} 
\usepackage{natbib}  
\usepackage{caption} 
\usepackage{algorithm}
\usepackage{algorithmic}
\usepackage{times}
\usepackage{helvet}
\usepackage{courier}
\usepackage{xcolor}
\usepackage{amsmath}
\usepackage{bm}
\usepackage{amssymb}
\usepackage{multirow} 
\usepackage{booktabs}
\usepackage{array}
\usepackage{amsfonts}
\usepackage{xcolor}
\usepackage{float}

\usepackage{newfloat}
\usepackage{listings}
\usepackage{array}    
\usepackage{booktabs} 
\DeclareCaptionStyle{ruled}{labelfont=normalfont,labelsep=colon,strut=off} 
\floatstyle{ruled}
\newfloat{listing}{tb}{lst}{}
\floatname{listing}{Listing}

\usepackage{cleveref}

\title{FantasyHSI: Video-Generation-Centric 4D Human Synthesis In Any Scene through A Graph-based Multi-Agent Framework}

\author{
    Lingzhou Mu\equalcontrib\textsuperscript{\rm 1,2},
    Qiang Wang\equalcontrib\textsuperscript{\rm 1},
    Fan Jiang\footnote{Project Leader}\textsuperscript{\rm 1},
    Mengchao Wang\textsuperscript{\rm 1},
    Yaqi Fan\textsuperscript{\rm 3},
    Mu Xu\textsuperscript{\rm 1},
    Kai Zhang\thanks{Corresponding Author}\textsuperscript{\rm 2}
}
\affiliations{
    \textsuperscript{\rm 1}AMAP, Alibaba Group\\
    \textsuperscript{\rm 2}Tsinghua University\\
    \textsuperscript{\rm 3}Beijing University of Posts and Telecommunications\\
    \textsuperscript{\rm 1}mulingzhou.m,yijing.wq,frank.jf,wangmengchao.wmc,xumu.xm@alibaba-inc.com \\
    \textsuperscript{\rm 2}mulz23@mails.tsinghua.edu.cn, zhangkai@sz.tsinghua.edu.cn \\
    \textsuperscript{\rm 3}yqfan@bupt.edu.cn
}

\usepackage{bibentry}
\begin{document}

\maketitle

\begin{abstract}
Human-Scene Interaction (HSI) seeks to generate realistic human behaviors within complex environments, yet it faces significant challenges in handling long-horizon, high-level tasks and generalizing to unseen scenes.  To address these limitations, we introduce FantasyHSI, a novel HSI framework centered on video generation and multi-agent systems that operates without paired data.  We model the complex interaction process as a dynamic directed graph, upon which we build a collaborative multi-agent system.  This system comprises a scene navigator agent for environmental perception and high-level path planning, and a planning agent that decomposes long-horizon goals into atomic actions.  Critically, we introduce a critic agent that establishes a closed-loop feedback mechanism by evaluating the deviation between generated actions and the planned path.  This allows for the dynamic correction of trajectory drifts caused by the stochasticity of the generative model, thereby ensuring long-term logical consistency.  To enhance the physical realism of the generated motions, we leverage Direct Preference Optimization (DPO) to train the action generator, significantly reducing artifacts such as limb distortion and foot-sliding. Extensive experiments on our custom SceneBench benchmark demonstrate that FantasyHSI significantly outperforms existing methods in terms of generalization, long-horizon task completion, and physical realism. Ours project page: \url{https://fantasy-amap.github.io/fantasy-hsi/}
\end{abstract}

\section{Introduction}

\begin{figure*}[h]
  \includegraphics[width=\linewidth]{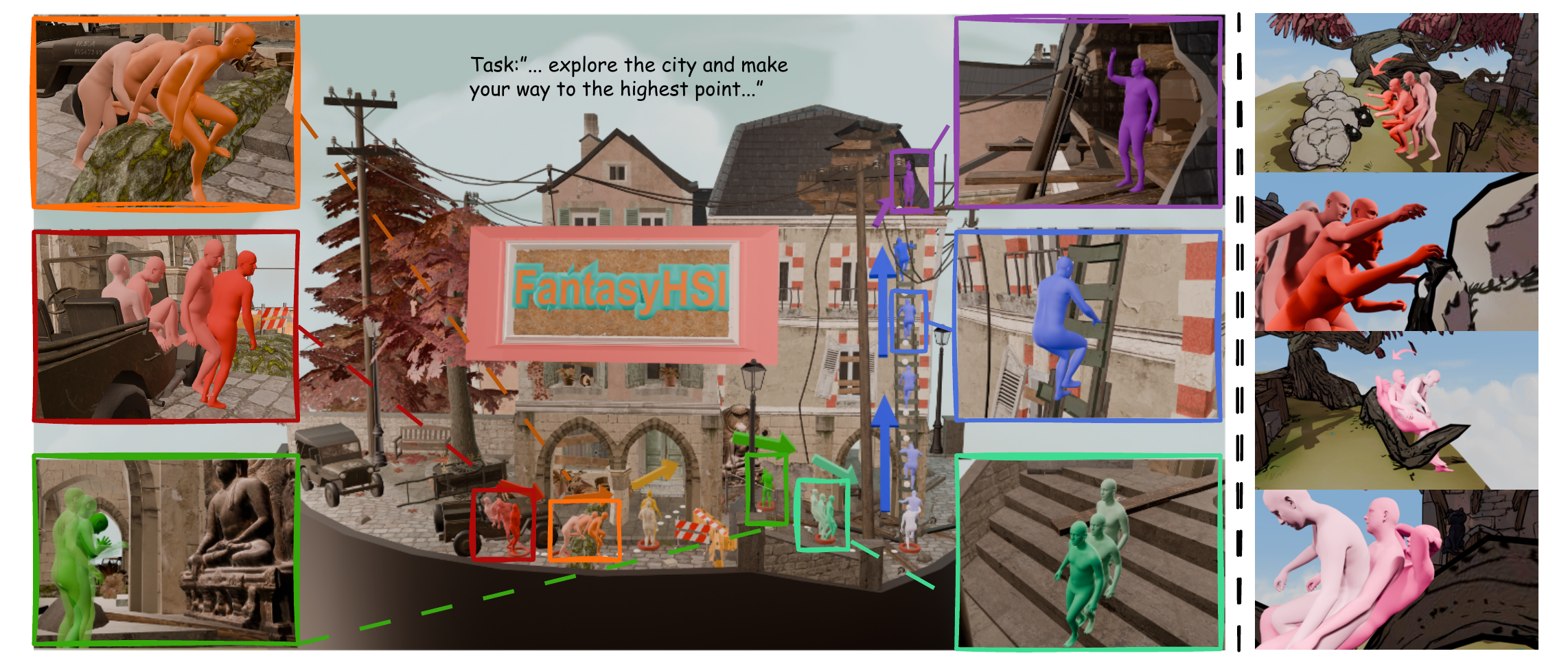}
  \caption{We introduce \textbf{FantasyHSI}, a novel framework that generates dynamic 4D sequences of humans interacting with their 3D environment. As illustrated on the left, FantasyHSI operates based on high-level task instruction, enabling it to autonomously plan paths, traverse obstacles, and execute a variety of complex motions, such as climbing a ladder. Moreover, the right side of the figure illustrates FantasyHSI's ability to generalize to arbitrary scenes and a variety of actions.}
  \label{fig:fig1}
\end{figure*}

Human-Scene Interaction (HSI) aims to understand and generate human movements in response to complex environmental contexts. This field has garnered increasing attention in computer vision and graphics due to its significant potential across diverse applications, such as embodied intelligence, virtual reality, and games.

As general intelligent agents, humans can perform a wide range of complex interactive tasks, flexibly respond to observed environmental information, and rapidly adapt to new surroundings. However, a significant gap remains between current methods and this level of human intelligence. Many approaches \cite{jiang2024scaling, cen2024generating, chen2024sitcom, jiang2024autonomous} rely on paired human-environment data, which typically requires collecting extensive matched motion capture and scene data within specific environments. Consequently, they lack adaptability when faced with unseen object layouts or dynamic changes, struggling to cover the rich diversity of real-world interactions. While some methods \cite{li2024zerohsi,li2024genzi} attempt to bypass the reliance on paired datasets by leveraging the prior knowledge of Vision-Language Models (VLMs) \cite{zhu2023minigpt, zhang2021vinvl,chen2023minigpt} or video diffusion models (VDMs) \cite{liu2024sora,wan2025wan,kong2024hunyuanvideo} to generate human-environment interaction sequences in a zero-shot manner, these are often limited to low-level, simple actions such as sitting or touching. They are ill-suited for high-level tasks, for instance, exploring a castle. Furthermore, generated motions must also be physically plausible. Any visual artifacts, such as limb deformation or foot-sliding, violate physical laws and severely diminish the realism and practical application of the results.

To address these challenges, we introduce FantasyHSI, a framework that models complex environmental scenes as a dynamic directed graph. By integrating VLM-based multi-agents with VDMs, FantasyHSI achieves effective environmental perception and planning, adjusts human motions based on environmental feedback, generates physically plausible human action sequences, and eliminates the dependency on paired human-environment datasets.

Specifically, we first introduce a dynamic directed graph to represent the human-environment interaction process. In this graph, nodes correspond to timestamped states of the human agent and the 3D scene, while edges encode the topological relationships of continuous action sequences. Building on this structure, we enhance the physical plausibility of VDMs using reinforcement learning, and leverage a motion capture system to acquire 4D spatio-temporal action sequences, which serve as the basis for dynamically generating the graph's edges.

Subsequently, we design a multi-agents system. This system features a scene navigation agent for environmental perception and understanding, and a planner agent that performs high-level task decomposition, breaking down long-horizon objectives into primitive actions. Critically, to address the inherent stochasticity of generative models, we introduce a critic agent to form a closed-loop feedback loop, which quantifies the discrepancy between generated actions and the planned trajectory, enabling dynamic correction of deviating node states.  This synergistic multi-agent architecture holistically unifies perception, planning, and correction, thereby resolving the issue of trajectory drift stemming from generative randomness and ensuring sustained logical coherence and physical viability in long-term interactions. Our primary contributions can be summarized as follows:

\begin{itemize}
    \item We propose a novel method of long-horizon human-environment interaction using a dynamic directed graph, which establishes an interpretable foundation for perception, planning, and behavioral refinement.
    \item We develop a collaborative multi-agent system that integrates environmental perception, path planning, and closed-loop correction to rectify action deviations caused by the inherent stochasticity of generative models.
    \item We design a controllable, physics-enhanced action generator by optimizing VDMs with reinforcement learning, which significantly improves the physical realism of the generated actions.
\end{itemize}

\begin{figure*}[ht]
    \centering
    \includegraphics[width=1.\textwidth]{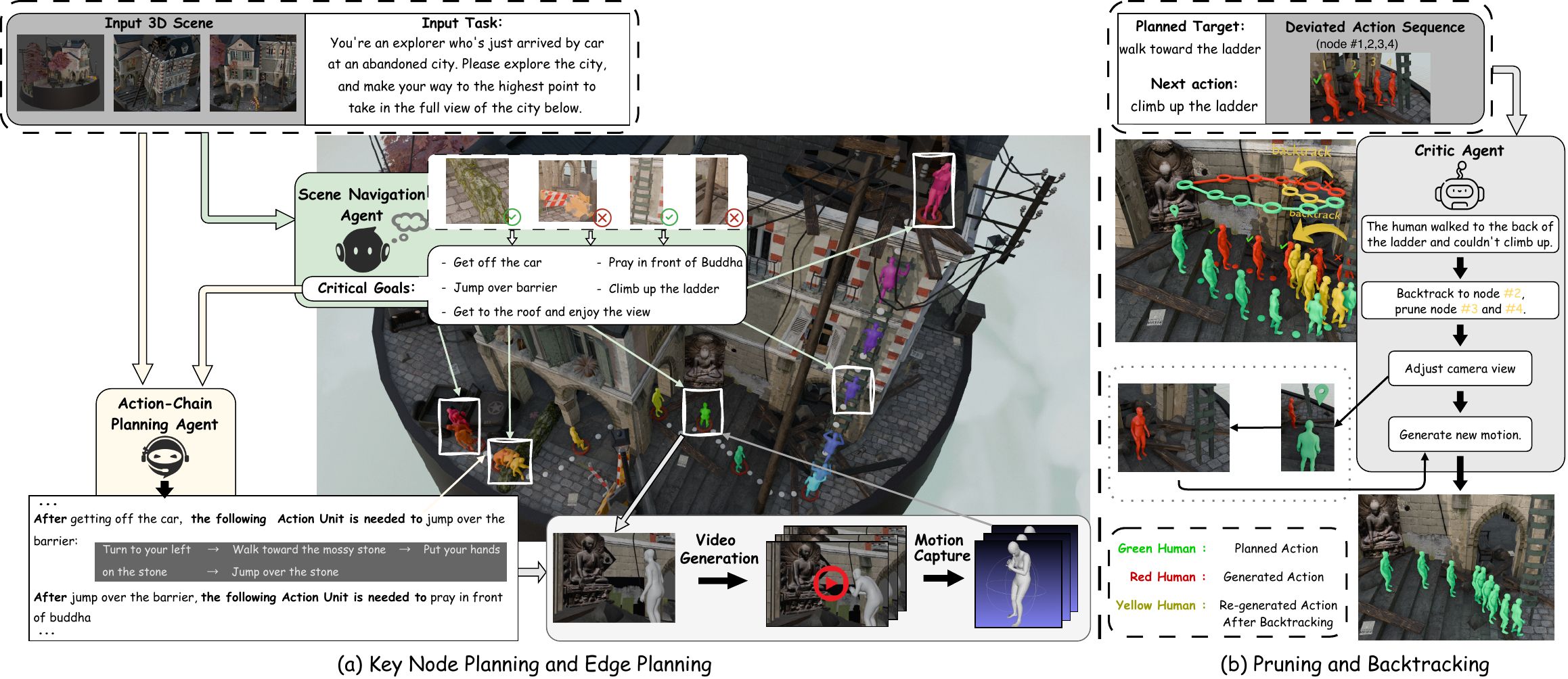}
    \caption{\textbf{Overview of FantasyHSI.}}
    \label{fig:overview}
\end{figure*}

\section{Related Work}

\subsection{Human-Scene Interaction Synthesis}
The objective of human-scene interaction synthesis is to generate realistic and coherent human motions that naturally interact with elements in a given environment. Early research \cite{tevet2022human, karunratanakul2023guided, chen2023executing} predominantly focus on modeling human behavior in isolation, neglecting the critical influence of contextual environmental factors. However, as human actions inherently constitute responses to external stimuli, recent approaches \cite{qing2023story, cen2024generating, chen2024sitcom, lou2024harmonizing} shift attention to scene-conditioned motion generation, incorporating action labels or textual descriptions as conditional inputs to enhance user controllability. By integrating environmental feedback and constraints, these methodologies achieve improved alignment with the dynamic nature of real-world interactions. Nevertheless, they typically rely on scarce paired action-scene data and exhibit limited generalization capabilities for novel interaction types. Although ZeroHSI \cite{li2024zerohsi} and GenZI \cite{li2024genzi} propose the zero-shot interaction generation paradigm that broadens the scope of motion synthesis applications, current implementations remain constrained to singular simple tasks, coupled with insufficient awareness of physical laws in generated outputs.

\subsection{Human Video Generation}
Human video generation from a single still image of a person, utilizing driving signals such as video \cite{xie2024x,wang2025fantasyportrait}, audio \cite{tian2024emo,cui2025hallo3,wang2025fantasytalking}, pose \cite{hu2024animate,zhu2024champ}, or text \cite{chen2024livephoto}. Early approaches \cite{prajwal2020lip,zhang2023sadtalker,ma2023dreamtalk} employ generative models like GANs \cite{goodfellow2020generative} or flow-based models \cite{ho2019flow++}. However, these methods often produce dynamic sequences suffering from artifacts and identity drift, resulting in insufficient realism and naturalness. Recently, diffusion models \cite{ho2020denoising} gain prominence in this field due to their demonstrated high quality and stability in image and video generation tasks \cite{rombach2022high,esser2024scaling,singer2022make}. Trained on large-scale datasets \cite{nan2024openvid,li2025openhumanvid}, diffusion models excel at modeling complex spatiotemporal relationships, yielding video results superior in visual quality and identity preservation. Nevertheless, current research \cite{wan2025wan,kong2024hunyuanvideo} primarily focuses on generating portrait videos in real-world settings. Their performance remains suboptimal when processing inputs like 3D scenes or SMPL-X \cite{pavlakos2019expressive} style images, and the generated videos often fail to strictly adhere to physical laws and exhibit deficiencies in character consistency.

\subsection{Reinforcement Learning from Human Feedback}
Reinforcement Learning from Human Feedback (RLHF) \cite{bai2022training} is a post-training methodology widely employed for large language models \cite{yuan2023rrhf,dubey2024llama,mehta2024openelm,stiennon2020learning} and diffusion models \cite{xu2024visionreward,clark2023directly,liu2025improving}, aiming to enhance model performance based on human feedback. Within this domain, Direct Preference Optimization (DPO) \cite{rafailov2023direct} is an effective method that trains the model using pairs of generated samples labeled as positive or negative. It enables the model to assign higher probabilities to preferred outputs and lower probabilities to less favored ones. By directly comparing different outputs generated by the model, DPO can more efficiently capture human preference information. DPO has been extensively applied to video diffusion models to improve visual quality, enhance motion coherence, and ensure spatio-temporal consistency, with its effectiveness validated in numerous studies \cite{liu2025videodpo,wu2025densedpo,wallace2024diffusion}. In our work, we employ the DPO approach to optimize video generation models, enabling them to produce content that better adheres to physical laws.

\begin{figure}[ht]
  \includegraphics[width=\linewidth]{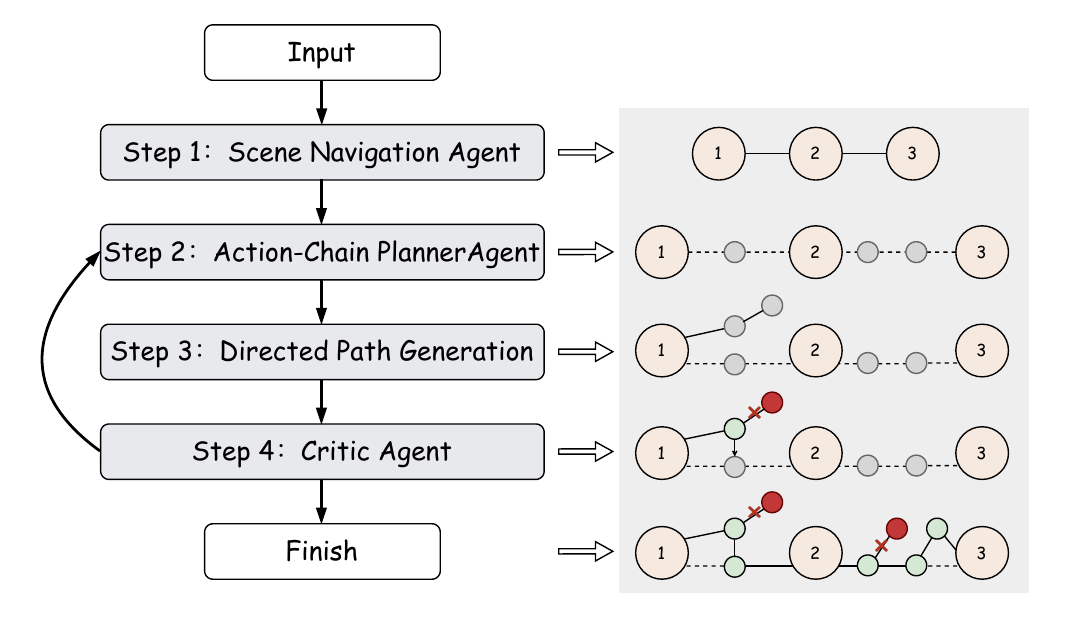}
  \caption{\textbf{The Multi-Agent Pipeline.}}
  \label{fig:dierzhang}
\end{figure}

\section{Method}
As shown in Figure \ref{fig:overview}, given the 3D scene and high-level instructions, we first formalize the task as a dynamic directed graph (Section \ref{sec:graph}), followed by task decomposition, planning, backtracking, and correction through a VLM-based multi-agents (Section \ref{sec:multi-agents}). We employ reinforcement learning to enhance the physical laws of the generator for each edge in the graph (Section \ref{sec: dpo}).

\subsection{Dynamic Directed Graph Representation}  \label{sec:graph}
\subsubsection{Graph Construction. }
To provide an interpretable representation for multi-agents, we model the task as a directed graph $G=(\mathcal{N},E)$, where $\mathcal{N}$ is the set of nodes and $E$ denotes the set of directed edges. Each node $N_k=\{\mathcal S_k, \mathcal H_k\} \in \mathcal{N}$ represents the state of the human and 3D scene at a specific time point. The human state $\mathcal{H}$ and the scene state $\mathcal{S}$ are represented by the 3D mesh. We employ SMPL-X \cite{pavlakos2019expressive} to model human pose and movement.

Furthermore, since some nodes represent the attainment of a critical goal, such as reaching the mountaintop or exiting the farm, we partition the node set $N$ into two categories: a set of key nodes $\mathcal{K}=\{K_1, K_2,...,K_p\} \subset \mathcal{N}$, which indicate signify milestone achievements, and a set of non-key nodes $\mathcal{U}=\{U_1,U_2,...,U_q\}=\mathcal{N} \setminus \mathcal{K}$, which correspond to the completion of individual action units but the critical goal remains unachieved.
Thus, a sequence of adjacent non-key nodes and corresponding edges can form a directed path that connects two key nodes, thereby representing the process by which a human character completes another critical task within the 3D scene via performing a series of  action units, culminating in the desired state denoted by the newly reached key node.

\subsubsection{Key Node Definition. }
A directed edge $E(N_i,N_j)=A_{i,j}$ indicates that the human $\mathcal{H}_i$ in node $N_i$ performs an action $A_{i,j}$, resulting in a new state described by node $N_j$. We formulate this state transition as $\mathcal{H}_j=\mathcal{H}_i + A_{i,j}$. Here, $A_{i,j}$ denotes an action unit that has complete semantic meaning. 

Given that human motion with long duration actually consists of continuous, frame-by-frame action sequences, naively defining human-scene state of each frame as an individual node would lead to excessive graph complexity. Thereby, only the state of the start and the end of an action unit $A_{i,j}$ is defined as node $N_i$ and $N_j$. In this manner, the human state before and after each action unit — $\mathcal{H}_i$ and $\mathcal{H}_j$ —together with their associated scene states $\mathcal{S}_i$ and  $\mathcal{S}_j$, constituted the adjacent node pair $N_i=\{\mathcal{H}_i,\mathcal{S}_i\}$ and $N_j=\{\mathcal{H}_j,\mathcal{S}_j\}$. 

\subsection{VLM-based Multi-Agents} \label{sec:multi-agents}
\subsubsection{Overview.}
As illustrated in Figure \ref{fig:dierzhang}, upon receiving a high-level task for the human in the scene, the scene navigation agent first analyzes the 3D scene and identify critical sub-goals required to accomplish the high-level task. Then it formulate a comprehensive plan that integrates both spatial trajectories and critical sub-goals to generate the key nodes $\mathcal{K}$ in the graph $G$. Subsequently, the action-chain planning agent generates a sequence of text described action chain $\{A_{i,i+1}, A_{i+1,i+2},...,A_{j-1,j}\}$, constructing a action chain connecting adjacent key nodes.
Next, the generator in Section \ref{sec: dpo} synthesizes human actions according to realize each action unit planned by the action-chain planning agent to construct a directed path. Due to the inherent randomness of the generative model, this process may introduce new, unplanned nodes into the original directed graph as the generated actions can differ from those initially planned. When such unplanned nodes arise, the critic agent analyses the nodes to route video generator return to the planned key node gradually when generating subsequent actions. This mechanism enables the agent to backtrack to wanted nodes, prune erroneous nodes, while refining viable ones. 

\subsubsection{Key Node Planning via Scene Navigation Agent.}
\label{sec: scene navigation agent}
The scene navigation agent is responsible for identifying critical sub-goals required to accomplish the given high-level task and generating a comprehensive plan that contains both trajectories and key events based on natural language description of high-level task $T$, the initial human position $\mathcal{H}_0$, and the initial 3D scene state $\mathcal{S}_0$. This plan is represented as a sequence of key nodes $\mathcal{K}$ in the graph $G$, as discussed in Section \ref{sec:graph}. 
The detailed reasoning process of the scene navigation agent is provided in Appendix A2.
 
\subsubsection{Edge Planning via Action-Chain Planner Agent.}
\label{sec: action chain planner agent}
Based on the key nodes and trajectory generated by the scene navigation agent, the action-chain planner agent decomposes the motion required to accomplish each sub-goals into a sequence of action units, each described in natural language. Here, an action unit is defined as a minimal semantic motion unit, representing a semantically coherent movement within three seconds. While complex human motions with long duration can be decomposed into short, meaningful action units, however, these atomic units are not not a finite set due to the vast variability of human motion in real-world scenarios. In our work, these action units may represent either complex human behaviors (e.g., ``yawning") or simple mechanical motions (e.g.,``turning backward").

While decomposing the motion into a sequence of more detailed action units, the action-chain planner agent actually extends the graph $G$ by adding intermediate non-key nodes $\mathcal{U}$ and a set of edges $E$ between adjacent key nodes $\mathcal{K}$. These added non-key nodes and edges can model state transitions between key nodes via a chain of action units, thereby bridging high-level planning with low-level motion execution.

\subsubsection{Directed Path Generation.}
\label{sec: video generation model}
The video generation model enhanced by physical laws in Section \ref{sec: dpo} serves as a human simulator in our framework, instantiate directed edges between all nodes to form a directed path that guides the agent from its initial state to each key goals planned by the scene navigation agent to accomplish the overall task. Our approach first generates videos for each action unit using a text-conditioned image-to-video model, then lifts the human motion into 3D scene via motion capture \cite{yin2024whac}. By iteratively rendering the final 3D state of the captured action and the scene as the initial frame for the next video generation, our approach can construct a continuous, scene-aware action chain, enabling the virtual agent to perform long-horizon, open-ended tasks in arbitrary environments.

Specifically, we first render a snapshot of the human $\mathcal{H}_i$ in scene $\mathcal{S}_i$ of the current node $N_i$ as the first frame of our video generation model, and use detailed description of the action unit $A_{i,j}$ in the graph as a prompt to generate a video clip. We then apply motion capture \cite{yin2024whac} to the generated video, extracting a 3D motion sequence in SMPL-X format. To lift the motion back to the 3D scene, we apply the motion change of each frame to the virtual human in the 3D scene. After completed the action unit $A_{i,j}$, the human-scene state is updated from node $N_i$ to $N_j$ through the instantiated edge $E(N_i,N_j)=A_{i,j}$. This newly updated node $N_j$ is then rendered as the next video’s initial frame, enabling iterative generation of subsequent actions. By repeating this process, we construct a directed path $\{N_0,E_{0,1},N_1,E_{1,2},...,N_k\}$ within the graph, enabling the virtual human to execute the actions and tasks in 3D space as planned by our agents.

\begin{table*}[ht]
    \centering
    \small
    \begin{tabular}{>{\centering\arraybackslash}p{2.50cm}|
                    >{\centering\arraybackslash}p{1.65cm}|
                    >{\centering\arraybackslash}p{1.25cm}|
                    >{\centering\arraybackslash}p{1.65cm}|
                    >{\centering\arraybackslash}p{1.65cm}|
                    >{\centering\arraybackslash}p{1.65cm}|
                    >{\centering\arraybackslash}p{1.25cm}|
                    >{\centering\arraybackslash}p{1.25cm}}

        \toprule
        \textbf{Method} & \textbf{P-Score}$\downarrow$ & \textbf{FS}$\downarrow$& \textbf{CLIP-S}$\uparrow$& \textbf{CLIP-C}$\uparrow$& \textbf{Diversity}$\uparrow$& \textbf{POS}$\downarrow$& \textbf{RDS}$\uparrow$\\
        \midrule 
        TRUMANS & 0.083 & 0.156 & 0.19 & \textbf{0.9943} & 0.1703 & 0.126 & 0.1806 \\
        LINGO & 0.085 & 0.148 & 0.20 & 0.9931 & 0.1510 & 0.133 & 0.1592\\
        PedGen & 0.063 & 0.199 & 0.17 & 0.9928 & 0.1102 & 0.119 & 0.1290\\
        Ours & \textbf{0.020}& \textbf{0.102} & \textbf{0.31} & \underline{0.9937} & 0.1911 & \textbf{0.063} & \textbf{0.3285} \\
        \midrule 
        SFT & 0.034 & 0.124 & \underline{0.27} & 0.9929 & 0.1221 & \underline{0.069} & \underline{0.3104}\\
        Baseline & 0.051 & 0.149 & 0.23 & 0.9865 & 0.1283 & 0.085 & 0.2173\\
        w/o multi-agents & 0.022 & \underline{0.103} & 0.24 & 0.9932 & \textbf{0.1920} & 0.070 & 0.2943\\
        w/o critic agent & \underline{0.021} & 0.105 & 0.23 & 0.9935 & \underline{0.1916} & 0.082 & 0.2891\\
        \bottomrule
    \end{tabular}
    \caption{\textbf{Quantitative Results of Comparison and Ablation Studies.}}
    \label{tab:ablation}
\end{table*}

\begin{figure*}[h]
    \centering
    \includegraphics[width=1.\textwidth]
    {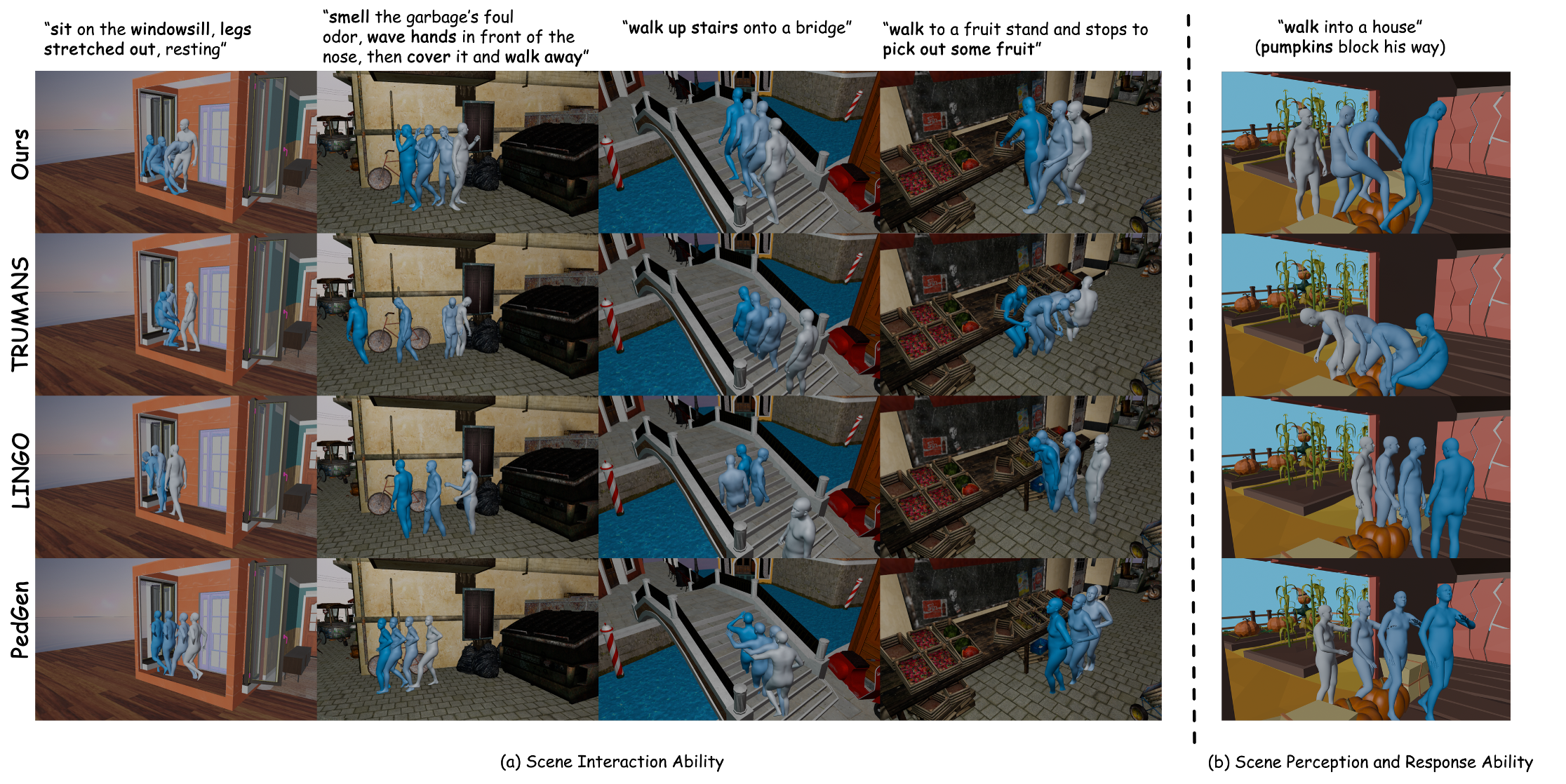}
    \caption{\textbf{Qualitative Comparison.}}
    \label{fig:cmp_fig}
\end{figure*}

\subsubsection{Pruning and Backtracking through Critic Agent.}
\label{sec: critic agent}
Due to inherent randomness in video generation process and the inevitable ambiguity of language prompt, the generated motion sequences can sometime lead the virtual human to deviate from the route planned by agents. 
For example, when generating a video clip of a person walking while enjoying the scenery, the distance traveled and the direction of movement can only be roughly controlled. Moreover, given an text described action like \textit{a sleepy stretch}, the video generation model might generate extra actions like an additional yawn following the stretch to express the person's sleepiness. Thus, this introduces new nodes into the graph constructed by the scene navigation agent and action-chain planner agent. In some cases, these extra actions and deviations enhance the expressiveness of the overall actions, but they might also be unwanted and disrupt the plan.

To process those newly generated nodes with deviations and unplanned actions, we utilize a critic agent that first assess these new nodes and then apply corrections if needed. Specifically, for each generated and captured motion segment, the critic agent analyzes the corresponding rendered frames, evaluates the motion, and corrects the trajectory and posture. Details of the evaluation and correction process can be found in Appendix A3.

\subsection{Physical Law Enhancement for Generator}
\label{sec: dpo}
We employ reinforcement learning to enhance the capabilities of video diffusion models in generating physically plausible motion and accurately following instructions.  Specifically, to improve performance in instruction following, motion artifacts (including clipping, unnatural motions), limb deformation, and scene inconsistency, we generate samples using four models: VEO \cite{deepmindveo2025}, HunYuan-Video \cite{kong2024hunyuanvideo}, Runway \cite{runwayml2025}, and Kling \cite{klingai2025}. Professional annotators then label positive samples $x^w$ and negative samples $x^l$ based on these criteria. We utilize DPO \cite{rafailov2023direct} to train the open-source Wan \cite{wan2025wan} model, thereby improving its ability to generate videos with enhanced physical realism and quality. Following DiffusionDPO \cite{wallace2024diffusion}, the training loss is defined as:

\begin{equation}
    \mathcal{L}_{\mathrm{DPO}} = - \mathbb{E}\Big[\log\sigma \Big(-\frac{\beta}{2}\big(L(x^w, p)-L(x^l, p)\big)\Big)\Big]
\label{eq:dpo}
\end{equation}
\noindent where $\beta$ is a temperature coefficient, $L(x^w , p)$ and $L(x^l, p)$ represent the losses for positive and negative parts, respectively. This loss encourages the generation of samples aligned with human preferences, promoting videos of human characters that are more physically plausible and consistent with reality.

\section{Experiment}

\subsection{Implementation Details} 
For video generator enhanced by physical laws using DPO, we utilize Wan2.1-I2V-14B \cite{wan2025wan} as base model. Then the model was trained on a collected dataset of 10,000 preference pairs for approximately 20 hours using 8 A100 GPUs, with a learning rate set to 1e-5. The $\beta$ is set to 5000. During the inference, we employed 30 inference steps and set the classifier-free guidance \cite{ho2022classifier} scale to 4.5. We employ Gemini-2.5-Pro \cite{GoogleDeepMindGeminiPro} as the VLM-based multi-agents.

\subsection{Evaluation Settings}
\subsubsection{Evaluate environmental Settings.}
To systematically evaluate our method, we conduct experiments under two settings, comprising scene interaction evaluation and scene perception and response evaluation.
The \textit{scene interaction evaluation} assesses the model’s ability to generate plausible human-scene interactions in static environments, where the scene geometry remains fixed throughout the motion. The \textit{scene perception and response evaluation} measures the model's ability to perceive and react to the changes and obstacles in the environment. In this setting, we introduce common real-world obstacle, both seen (e.g., chairs, sofas, vases) and novel (e.g., pumpkins, rocks), into the model’s pre-planned path. The model must first detect the obstacle and then react to it accordingly. This evaluates not only robustness to unseen objects but more importantly evaluates how the model perceive the world and react to the world.

\subsubsection{Evaluation Dataset.}
Due to the lack of publicly available HSI benchmarks, systematic evaluation remains challenging. For instance, TRUMANS \cite{jiang2024autonomous} only released its training dataset without a standardized test set, and other works such as LINGO \cite{jiang2024autonomous} have not yet made their evaluation sets publicly available. Therefore, we introduce SceneBench, an evaluation benchmark comprising diverse 3D environments designed to assess embodied virtual human behavior in indoor and outdoor scene. The scenes are sourced from TRUMANS, and our web-collected scenes from Sketchfab \cite{sketchfab}, resulting in a total of 20 distinct 3D scene, 10 indoor and 10 outdoor, spanning residential spaces (e.g., bedrooms, cowshed, gym), natural landscapes (e.g., grasslands, riversides), urban streets, and rural farms, etc.

For scene interaction evaluation, each scene is annotated with human-scene interaction tasks described in natural language, along with their start and goal positions. This yields 120 text-scene-position pairs as the test instances for evaluation. For scene perception and response evaluation, we sample 15 objects from SceneBench and 15 additional 3D models collected from the Internet. These include both seen household objects from TRUMANS and unseen, out-of-distribution objects. We place them along pre-planned paths to evaluate how models react to them. (See Appendix B1. for more details.)

\subsubsection{Baselines.}
We compare our method with recent HSI methods, as well as a scene-aware 4D pedestrian generation method, including TRUMANS \cite{jiang2024scaling}, LINGO \cite{jiang2024autonomous} and PedGen \cite{liu2024learning}. 

Unlike baseline methods that require predefined path specifications, our multi-agents system operates end-to-end, generating both path and motion plans directly from higher-level natural language instructions. Specifically TRUMANS needs a full trajectory input, while LINGO and PedGen requires start and goal positions. In our experiments, we provide these additional input using ground-truth locations to adapt the baseline methods to our task and thus fully demonstrate the capabilities of the baseline methods.

\subsubsection{Metrics.}
For evaluation, we adopt metrics from prior works \cite{jiang2024scaling, liu2024learning, li2024zerohsi} to assess both scene interaction and motion quality. To evaluate scene interaction, we employ the Penetration Score (\textbf{P-Score}) \cite{zhao2023synthesizing} to measure the percentage of body vertices penetrating the scene, and the Foot Sliding Score (\textbf{FS}) \cite{he2022nemf} to quantify foot sliding. To assess motion quality and diversity, we utilize the \textbf{CLIP-Score} \cite{radford2021learning} for measuring the alignment between the generated motion and the input text prompt, alongside \textbf{CLIP-Consistency} to evaluate frame-wise temporal coherence, motion \textbf{Diversity} to evaluate diversity under same inputs. In addition, to evaluate scene awareness, we introduce two metrics. The Penetration Obstacle Score (\textbf{POS}) measures mesh penetration between the human and introduced obstacles, while the Reaction Divergence Score (\textbf{RDS}) computes the average per-joint distance between motions generated with and without obstacles.

\subsection{Evaluation results}
\subsubsection{Scene Interaction Ability Evaluation.}

As shown in Figure \ref{fig:cmp_fig}(a), we present qualitative comparison of scene interaction ability between FantasyHSI and baseline approaches on SceneBench. Our results demonstrate that our method generates vivid and expressive motions in various environments, accomplishing diverse high-level human-scene interactions beyond simple walking or touching. For instance, our method can produces highly abstract and human-like behaviors such as fanning one’s nose near a garbage pile, sitting on unusual places like windowsills, and even climb 20 meters up the ladder to reach the rooftop in Figure \ref{fig:fig1}, while all of the other methods failed in these tasks. 

Qualitative results indicate that TRUMANS is severely overfitted to its training distribution, as it defaults to generating only sitting motions when encountering novel objects. As shown in the first column of Figure \ref{fig:cmp_fig}(a), it fails to perceive the windowsill's height, instead generating sitting poses at a standard chair height consistent with its training data. Furthermore, LINGO struggles to perceive surface boundaries in unseen environments, such as the third and fourth columns in Figure \ref{fig:cmp_fig}(a), and present limited scene understanding. As a result, it suffers severe penetration and are unable to produce reasonable motions for highly abstract interaction tasks in column 2. Although PedGen generates temporally coherent walking sequences, its motion diversity is highly limited and lacks the capability to perform meaningful scene interactions.
As evidenced in Table \ref{tab:ablation}, our method achieves the highest CLIP Score and diversity with the lowest penetration and FS, outperforming existing approaches across most metrics. This demonstrates that our method generates more semantically aligned and physically plausible motions with better diversity.

\subsubsection{Scene Perception and Response Ability Evaluation.}
For the evaluation of scene perception and response capability, we present qualitative comparison results in Figure \ref{fig:cmp_fig}(b). Among all these methods, only our approach successfully perceived the obstacles (the pumpkins) and generated reasonable reactive behaviors, such as stepping over them. Although TRUMANS and LINGO detect the presence of obstacles via occupancy grids, their perception range is limited to a 1-meter box around the virtual human, represented by point clouds. The point clouds of surrounding objects are truncated by this limited sensing range, preventing the models from perceiving the full structure of the objects and leading to significant loss of semantic information. As a result, LINGO generates a backward glance, while TRUMANS fail to generate reasonable motions, neither sucsessfully avoids or interacts with the obstacle. In contrast, PedGen demonstrates very poor obstacle awareness, simply walking through the pumpkins without any reactive behavior. Align with visual observations, as shown in Table \ref{tab:ablation}, our method also outperforms all methods on Penetration Obstacle Score and Reaction Divergence Score, indicating superior scene understanding and response ability.

\subsection{Ablation Studies and Discussion}

\begin{figure}[ht]
  \includegraphics[width=\linewidth]{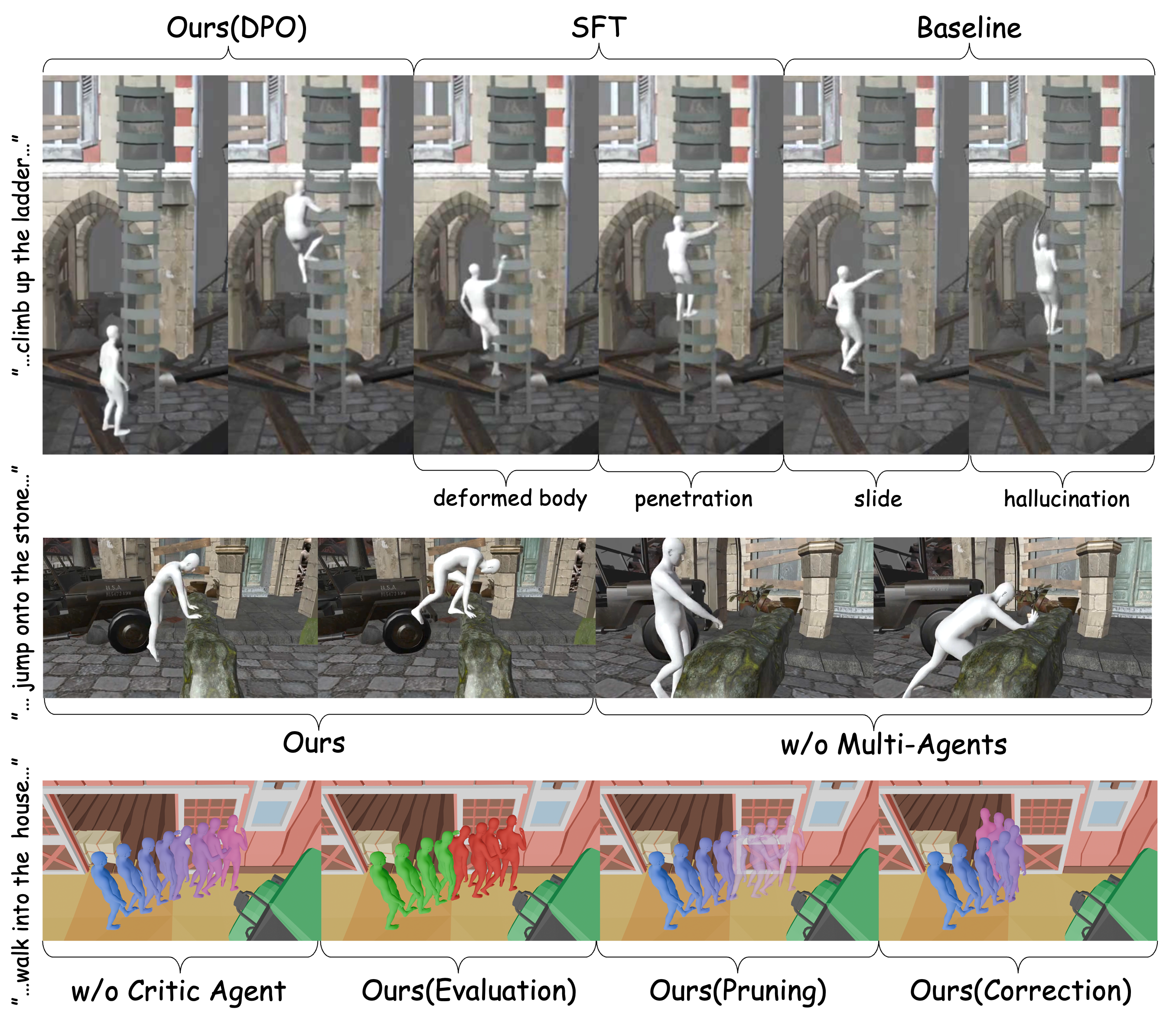}
  \caption{\textbf{Qualitative Results of Ablation Studies.}}
  \label{fig:ablation}
\end{figure}

\subsubsection{Multi-Agents. } 
To evaluate the effectiveness of our multi-agent collaborative framework, we conducted an ablation study in which no agent was involved in neither planning the actions or decomposing complex actions into a chain of action units. In this setting, complex motions were generated directly using video generation model. As shown in the second row of Figure \ref{fig:ablation}, when prompted to jump on to the fence, the model fails to generate desired actions without multi-agents, since it lacks of detailed motion planning as instructions.By contrast, our approach break down the complex motion into a chain of Action Units. With this detailed plan, the virtual human is first instructed to places his hands on the rock for support, and then the he jumps up and lands with both feet on top of the rock, thereby successfully completing the overall action. Additionally, as shown in Table \ref{tab:ablation}, the significant drop in the CLIP-S score indicates that our method fails to accomplish the task objective without the multi-agent component to decompose the main goal into clear sub-tasks. This result validates the effectiveness of multi-agents.

\subsubsection{Critic Agent.}
To evaluate the effectiveness of the critic agent in our method, we conducted ablation studies comparing results with and without its involvement. As shown in Figure \ref{fig:ablation}, without the evaluation and backtracking mechanisms provided by the critic agent, the model is unable to correct deviations from the intended path, ultimately failing to reach the planned target location. In contrast, when the critic agent is incorporated, the model successfully re-navigates the virtual human to the target position. Furthermore, as evidenced by Table \ref{tab:ablation}, the absence of the critic agent causes a significant drop in CLIP score, indicating that the model struggles to accomplish the specified goals. The observed increase in the Diversity metric is primarily due to the inclusion of deviated motion segments that would otherwise be backtracked and pruned by the critic agent.

\subsubsection{DPO for Video Generative Model.}
To validate the effectiveness of our video generation model optimized with DPO, we conducted comparative experiments on the test set using both the Supervised Fine-Tuning (SFT) model and the original pre-trained model. 
As shown in Figure \ref{fig:ablation} and Table \ref{tab:ablation}, while the base model and the SFT method demonstrate a certain degree of instruction-following capability, their generated outputs often exhibit physically implausible dynamics. These include artifacts such as characters penetrating the environment, body distortions, and unnatural sliding motions. In contrast, our method, optimized with the DPO, significantly enhances the ability to generate dynamics consistent with real-world physics, thereby achieving superior results.

\subsubsection{Limitation and Future Works.}
Notwithstanding the notable advancements of FantasyHSI in addressing complex human-scene interactions and long-horizon tasks, certain limitations persist that merit further exploration. A primary concern is the limited inference capabilities of current diffusion-based video generation models and VLM-based agents. This computational bottleneck potentially hinders the deployment of our method in real-time, interactive settings.  Furthermore, due to the scarcity of dynamic environment datasets, our research has focused on static environments, which deviates from the dynamic nature of real-world scenarios. Therefore, a key direction for future research is to extend our capabilities to more practical and realistic dynamic environments.

\section{Conclusion}
In this work, we presented FantasyHSI, a novel framework for synthesizing expressive and physically plausible human-scene interactions in complex 3D environments. By reformulating HSI as a dynamic directed graph, we established an interpretable structure for modeling long-horizon interactions. The integrated VLM-based multi-agent collaboration comprise scene understanding, hierarchical planning, and trajectory correction. Furthermore, our reinforcement learning-based optimization of video diffusion models ensures that synthesized motions adhere to physical laws, eliminating artifacts such as foot sliding and body-scene penetration. Experiments show that FantasyHSI surpasses existing methods in generalization to unseen scenes and long-horizon tasks while maintaining motion realism and logical coherence.

\bibliography{aaai2026}

\newpage

\begin{figure*}[h!]
  \includegraphics[width=\linewidth]{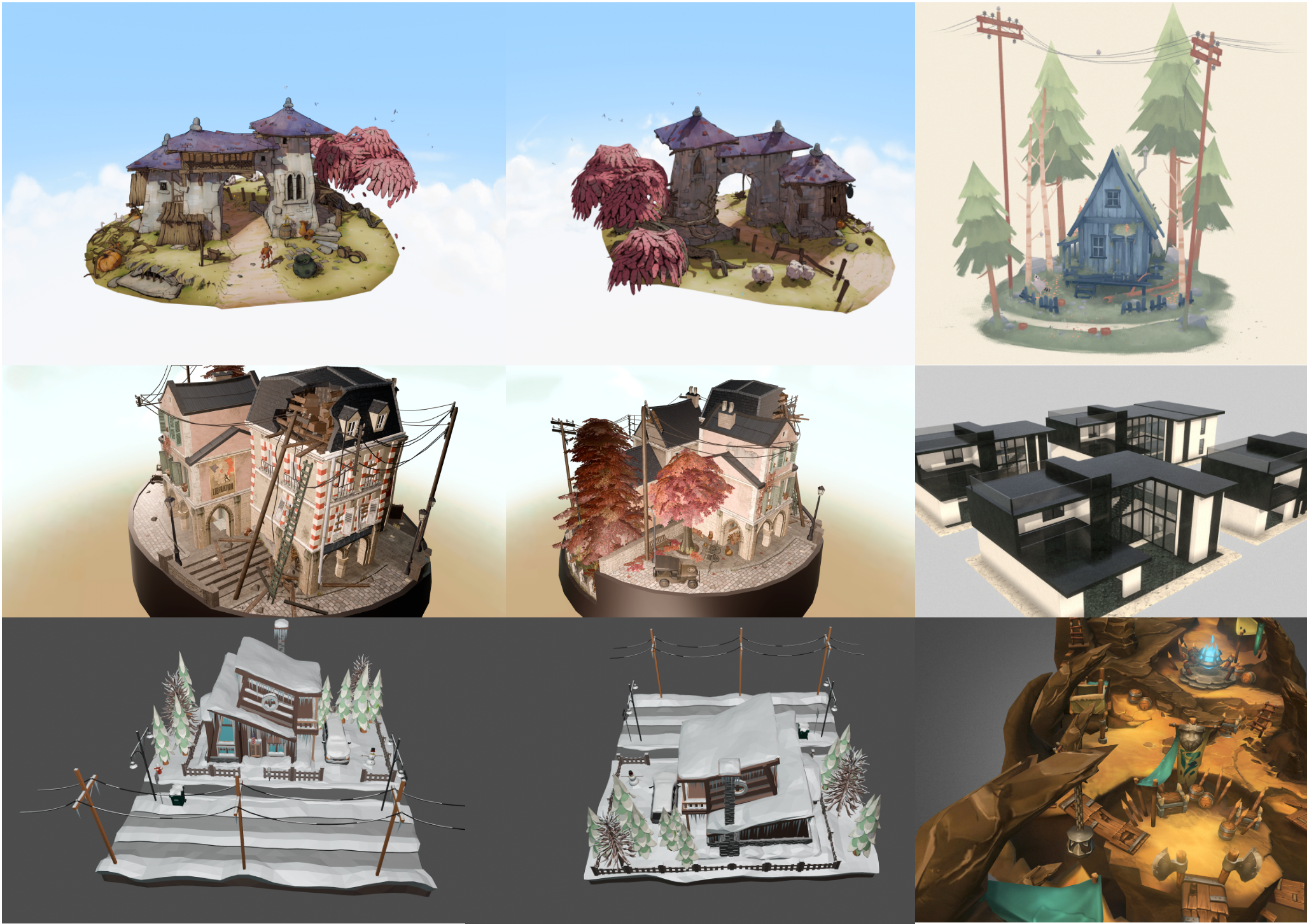}
  \caption{Example of scenes in our SceneBench.}
  \label{fig:scenebench}
\end{figure*}

\section{Supplementary materials}

\subsection{A. Method}
\subsubsection{A1. Establishment of Human State}

The human state in our framework is defined as $\mathcal{H}=F(R,T,\theta,\beta)$, capturing the intricacies of human representation through a set of parameters. Here, $R$ and $T$ refer to the rotation and translation vectors, respectively, which are fundamental for positioning and orienting the human mesh in a 3D space. The parameter $\theta$ embodies the human pose, detailing the angles and orientations of various joints, and is crucial for realistic depiction of posture and movement. Meanwhile, $\beta$ denotes the body shape parameters, influencing the overall physique and proportions of the 3D model to match individual characteristics. The function F serves as the SMPL-X \cite{yin2024whac} rendering function, which generates the 3D human mesh given these parameters.

\subsubsection{A2. Reasoning process of the Scene Navigation Agent} 
The Scene Navigation Agent follows a structured reasoning process: 
\begin{enumerate}
    \item \textit{Semantic Analyses of the 3D Scene}: The VLM-based agent first sees several top-down view of the 3D scene, identifying environmental elements and their spatial relations. 
    \item \textit{Accessible Space Identification}: based on scene understanding, the agent distinguishes between navigable pathways, accessible regions, and non-traversable obstacles to avoids routing the human into inaccessible areas. 
    \item \textit{Intent recognition}: the agent analyses the input task description $T$ and the environment to infer the behavioral intentions and preferences of the human. 
    \item \textit{Interactive Objects Identification}: The agent identifies objects within the scene that afford interaction (e.g., chairs, cars, animals) 
    \item \textit{Sub-goal and Path Planning}: Integrating the outputs from the preceding reasoning steps, the agent formulates a sequence of sub-goals and an associated navigable path. This process generates a sequence of key nodes $\mathcal{K}$, which initializes the graph $G$ and describes a trajectory within the environment to accomplish the given task.
\end{enumerate}

\subsubsection{A3. Details of Critic Agent}
For each generated and captured motion segment, the Critic Agent analyzes the corresponding rendered frames, assess the motion and performs the rectification via following steps:
\begin{enumerate}
    \item \textit{Pose, Distance, and Trajectory Evaluation}: The agent sees several rendered frames and evaluates the virtual human’s proximity to the targeting place of this action, along with its orientation toward the goal, and adherence to the planned route.
    \item \textit{Temporal Backtracking}: Given the generated $n$ frame motions, the agent starts from the final frame, and backtracks the motion every few frames, identifying the latest frame $i$ that best matches the expected motion in terms of action semantics, physical plausibility and spatial position.
    \item \textit{Graph Pruning}: The actions from the $i^\textit{th}$ frame to $n^\textit{th}$ frame are then discarded as well as the corresponding edges (actions) and nodes the graph $G$ are pruned.
    \item Spatial Orientation Correction: If the agent’s orientation in the $i^\textit{th}$ frame does not align with the target direction, the Critic predicts a corrective turning angle and smoothly distributes the rotation across the remaining $i$ frames ensuring a natural transition.
    \item \textit{Camera Pose Adjustment}: Since we require fixed-camera video generation to avoid hallucination of unobserved regions, the initial frame’s camera pose is critical. Thus, after steps 1–4, the VLM based agent sets an optimal camera viewpoint in the 3D scene to render a snapshot as next initial frame, ensuring high-quality video generation. The camera must clearly capture the target region without occlusion, with the human centered and appropriately scaled.
    \item \textit{Future Plan Adjustment}: If backtracking and pruning is applied, the critic agent then adjusts the plan made by previous agents to guide the virtual human back to the original planned route in the next few video clips. 
\end{enumerate}

\subsection{B. Experiment}
\subsubsection{B1. Visualization of our SceneBench}
As shown in Figure \ref{fig:scenebench}, we present several examples from our scene dataset. It can be observed that our scenes exhibit a high degree of diversity, including real scenes, abstract-style scenes, stylized scenes, and scenes with complex 3D spatial structures.

\end{document}